# Role-Aware Multi-modal federated learning system for detecting phishing webpages

**Bo Wang *, Dr. Imran Khan, Prof. Martin White, and Dr. Natalia Beloff**

University of Sussex, Brighton, United Kingdom
*Correspondence: author: bw268@sussex.ac.uk*

**Abstract.** In this paper, we propose a multi-modal phishing website detection system based on federated learning, which can simultaneously utilise URL, HTML, and IMAGE data, and does not require the client to be bound to the data type during the inference stage: any client can invoke the corresponding modality head trained by other clients for detection. Methodologically, we implement a bucket aggregation strategy by role (expert level) on the FedProx paradigm, drawing on the multi-expert idea of Mixture-of-Experts (MoE) and the design concept of the FedMM algorithm. We remove the learnable routing and adopt hard gating (directly selecting the IMAGE/HTML/URL expert based on the modality label of the sample) to separately aggregate the parameters of each modality, to isolate the aggregation conflicts and convergence oscillations caused by different embeddings. We validate the effectiveness on two public datasets (TR-OP and WebPhish): the Fusion head achieves performance metrics of Accuracy (Acc) 97.5%, Precision (Prec) 98.1%, Recall (Rec) 97.3%, and 2.4% FPR on the two types of data on TR-OP. Meanwhile, in the ablation experiment, it achieves Acc 95.5% / Prec 94.3% / Rec 96.5% / FPR 5.9% on the image subset of TR-OP. On the text side, we applied two adaptations: for the simple-structured URL, we directly use GraphCodeBERT to generate embeddings; for the raw and uncleaned HTML, we adopt an early fusion of a three-way embedding design to reduce noise. On WebPhish's HTML (relatively simple), we achieve Acc 96.5% / Prec 98.1% / Rec 95% / FPR 1.8%; on the raw HTML of TR-OP (relatively difficult), we achieve Acc 95.1% / Prec 95.4% / Rec 94.6% / FPR 4.6%. The results show that bucket aggregation with hard-gating experts can stably conduct federated training and enhance the usability of multi-modal under strict privacy boundaries, providing a more flexible and comprehensive federated solution for phishing website detection.

# 1 Introduction and Related Work

## 1.1 Development of phishing webpage detection

Phishing webpages, as a persistent problem in web security, have attracted continuous attention from researchers due to their variability, complexity, and the constant emergence of new web technologies and third-party development platforms [17][33]. Phishing webpages tend to conduct deceptive attacks on users, tricking them into inadvertently disclosing their personal privacy information and security data, resulting in significant financial losses. Traditional methods typically rely on blacklists to directly compare the URLs or domains of the webpages that users visit [36][37], and warn or block the corresponding pages before users start entering any sensitive information. However, this manual rule-based approach, while ensuring a certain degree of interpretability, is unable to handle unseen page or URL types. Therefore, machine learning-based detection solutions have emerged [1][4][8]. These solutions offer a more automated and efficient approach, eliminating the need for extensive manual rule updates. In recent years, with the rise of large models, there has been an increasing number of studies exploring the use of large models to replace traditional machine learning end-to-end solutions, further enhancing the generalisation and coverage of machine learning-based approaches.

## 1.2 Existing Phishing Webpages Detection Techniques

Based on our investigation of the literature, the research on the automatic detection of phishing webpages and machine learning-related techniques in recent years can be roughly divided into three major categories: 1. End-to-end models based on machine learning [1][4][8][16][17][18]; 2. Reference-based models that combine machine learning or LLMs [5][6][7][11][12]; 3. Direct detection of phishing webpages using LLMs [2][10][14].

The end-to-end detection of phishing webpages can be roughly classified into these categories based on the way data is utilised: 1. Detection of web source code using features of HTML or URL, or a combination of both; 2. Detection using images and logos, or a combination of their features; 3. Selection of an associated combination among HTML, URL, and images, followed by feature fusion and classification. Each of these studies has its own advantages and disadvantages (mainly in the trade-off between computational load and performance), such as the relatively early studies HTMLphish [4] and Web2Vec [16], which achieved relatively comprehensive utilisation of data like HTML and URL. Particularly, Web2Vec treated URL, page text content, and DOM structure as sequences (character/word/sentence granularity) for learning representations, followed by multi-channel feature extraction and fusion, and finally binary classification. Each channel went through CNN→BiLSTM→Attention (CNN captures local features, BiLSTM captures context, and attention emphasises key information). Their final performance (99% accuracy, 0.25% FPR) demonstrated the effectiveness of this fusion strategy, but it also greatly increased the computational load.

WebPhish [8] is another relatively lightweight case in recent years. They proposed to concatenate HTML and URL features into a complete vector and then feed it into a downstream classifier, similar to the earlier Web2Vec approach. This scheme effectively utilised both types of data through early fusion and achieved very satisfactory performance metrics (98% accuracy and recall). Their dataset was quite large (45,373 samples, balanced positive and negative examples), covering most of the features of phishing and normal pages to a high degree. Although they did not use any image data and the training method was traditional centralised, it was still a good design that utilised both types of data simultaneously.

The approach of using the MoE (mix of experts) concept for complex design also exists. In the research by Wang's team [1], they decomposed the URL into attributes: WHOIS for the duration of domain registration; content: whether there is a privacy request/pop-up/mask; behaviour: the number of redirects and loading delay (normalised using the mean of the top 100 websites). These statistics are uniformly encoded into vectors of the same dimension as the token embeddings (binary encoding of registration duration → 756 dimensions), and fed into a fusion model of Transformer + Mixture-of-Experts. In their tests, MoE did indeed improve the performance decline caused by the small amount of training set data during centralised training. Although there was not much improvement in pure classification performance, and the data was limited to URLs, the overall design idea was indeed very rare.

Directly using LLM (Large Language Model) for phishing detection is also a direction [2][10][14]. Although there are many problems with this approach, such as the unexplainability of LLM, weak deployment flexibility (usually only API calls can be made and HTML consumes a lot of costs), and the need to separately test the resistance to adversarial attacks and prompt attacks, the powerful performance of LLM (such as ChatGPT) still brings visible performance improvements to phishing detection. For example, in Lee's team's research, they conducted a two-stage test based on brands for three large models. In the first stage, they used a multimodal LLM to identify the target brand from screenshots and/or simplified HTML. In the second stage, they used LLM to verify the consistency between the brand and the domain name. They achieved relatively good performance indicators on their self-built dataset. Moreover, LLM can achieve better results than single visual models that only use logos in noisy environments by leveraging multimodality.

Regarding research on hybrid machine learning models and reference-based schemes, such studies mainly utilise the explainability of reference-based methods and the efficiency of machine learning to achieve higher performance and more explainable detection results. At the same time, because these schemes are usually more modular, they are more friendly to multimodal design (allowing individual modules to handle specific modalities). Early well-known examples include PhishPedia [12] and, recently, Phishagent [11].

The early hybrid phishing detection framework, PhishPedia, is a very famous study. They used brand logos + input boxes + target domains. They rendered page screenshots in a sandbox based on the URL, then used Fast-RCNN to detect logos and input boxes, followed by using Siamese for comparison between logos and the knowledge base (static), identified the brand, and compared it with the official brand library to decide. At the same time, they also emphasised the common problem in phishing detection using machine learning paradigms, that is, over-reliance on a single model may cause the model to learn the correlation between datasets rather than the features of the data itself, leading to a high FPR rate. We hold a reserved view on this point because many current machine learning-based phishing detection schemes indeed have significant differences in datasets and comparisons (many methods claiming to be SOTA achieve around 99% performance on their own datasets), resulting in unfair performance comparisons.

Meanwhile, we agree that the FPR of a model is a very important performance indicator, so we specifically observed this issue in our experiments. PhishPedia, being an early study, has been surpassed in recent research, such as Phishagent (which includes performance comparisons on the same dataset). The recent hybrid design of Phishagent proposes to use a large model combined with a hybrid brand knowledge base to achieve a more comprehensive multimodal detection of phishing websites. In their design, the large model plays a key role, similar to that in ChatPhish [2]. Unlike ChatPhish, they break down the entire multimodal system into more detailed sub-modules (offline knowledge base module, online knowledge base module, core decision agent) and split the web page into three types of inputs: W.HTML, W.screenshot, W.URL, and integrate them into a single input W (webpage). They complete the retrieval and discrimination within one iteration, balancing low latency and brand coverage. However, their ablation experiments also exposed some problems, that is, they are highly dependent on the online query module. Once this module is removed, the performance drops sharply, which is a common problem for some solutions that rely on rules and knowledge base queries, such as PhishPedia.

### 1.3 Federated learning and phishing webpage detection

Currently, research directly applying federated learning to phishing webpage detection is scarce or very limited. Most studies mainly focus on the application of federated learning to phishing emails, text messages, or URLs [3][15][20][21]. The attack surfaces targeted by most of these studies are either unrelated to phishing webpages or relatively narrow. For instance, STFL [15] proposed using federated learning to detect phishing URLs. They utilised three datasets to enhance coverage of URL datasets and classified URLs into three feature categories: domain + TLD / subdomain/path. They performed one-hot encoding on characters and selected 52 high-frequency characters, then concatenated them. However, their overall approach was limited to URL data alone and did not discuss other features of phishing webpages, such as HTML itself. In their explanation, they chose to skip the analysis of web content (such as HTML) due to the complex computations and significant time requirements involved.

Moreover, although some research has considered the data privacy issues of phishing webpages, such studies are still extremely rare. Early on, the Microsoft team proposed moving phishing webpage data to the cloud to prevent the leakage of sensitive information (such as email addresses, bank details, and corporate secrets) contained in the data [19]. They used a third-party platform model to train and detect without accessing the original phishing pages. STFL also mentioned that URLs, as one of the features, might contain users' personal information, making centralised learning inappropriate, especially considering legal requirements like GDPR. These requirements are highly consistent with the principles advocated by federated learning, such as data localisation.

One of the biggest advantages of federated learning is the localisation of data. Although this may lead to some performance loss, studies have shown that due to the heterogeneous and non-IID (Non-independent and non-identically distributed) nature of federated learning, it is difficult to directly compare its performance with centralised learning [23][24]. Of course, considering its privacy protection and data localisation advantages, this can be seen as a trade-off.

### 1.4 Contributions

Based on all the research mentioned above, considering the scarcity of current studies on phishing detection using federated learning, especially in the multi-modal aspect, and the deviation in practical usability caused by the one-sided application of data in end-to-end design, we propose a federated learning-based multi-modal phishing detection system to address the following challenges, which are also the contributions of this paper:

- **The diversity and complexity of phishing website pages:** It is difficult to reflect the complexity of phishing websites by using only a single type of data. We aim to, through the concept of MoE (though not strictly MoE), enable the global model to handle multiple types of data (IMAGE, HTML, URL) simultaneously under federated learning, and still achieve acceptable detection results when a certain type of data is missing.
- **Data dependency:** We do not require the test sets and training sets of clients to be equivalent or strongly dependent. We decouple the relationship between the test set and the training set on the client side. Any client can use the corresponding modality heads trained by other clients for detection, as long as the data type is allowed by the framework.
- **Data privacy:** The inherent nature of phishing websites existing in direct user interaction may lead to the capture of some users' personal information, making the original data unable to be shared. Given the poor commonality of the previously mentioned datasets, we hope to effectively utilise multi-source and multi-party data through federated learning, ensuring that the original data does not circulate and does not violate relevant laws.

- **Aggregation conflicts in federated learning with multi-modalities:** It is difficult to aggregate multiple types of data that are fundamentally different using federated learning alone, as there are fundamental differences in embedding and semantic spaces, which can cause conflicts during model aggregation. If separate federations are conducted for each type of data using a single model, it will greatly increase computational and time costs. We design a role-based aggregation strategy, where each client only aggregates the corresponding data, to achieve isolation of aggregation parameters and multi-modal learning in a single federation.

# 2 Methodology

## 2.1 General framework design

In this paper, we made use of a federated learning simulation system based on Flower [28]. Flower is an open-source federated learning simulation framework which supports a variety of mainstream algorithms and highly flexible custom aggregation strategies.

We designed four distinct expert module routes to process three different types of data, namely pure IMAGE headers, pure HTML headers, URL headers, and a Fusion header that combines IMAGE and HTML. Given that the difficulty levels of different tasks vary, the corresponding pre - processing, embedding schemes, and downstream classifiers also differ. The three types of data are HTML, IMAGE, and URL. Judging from the difficulty, URL is the least complex, followed by IMAGE. HTML is the most difficult because it involves different and complex HTML elements from two datasets simultaneously. In the section on dataset explanation, we elaborate on the reasons for the substantial differences in the difficulty levels of these three types of data.

We adopted three different embedding methods: Pix2Struct [13], GraphcodeBERT [22], and static three-way segmentation (the process is elaborated in detail in the data pre - processing section), similar to the approach of WebPhish. The overall process of the entire framework is presented in Figure 1.

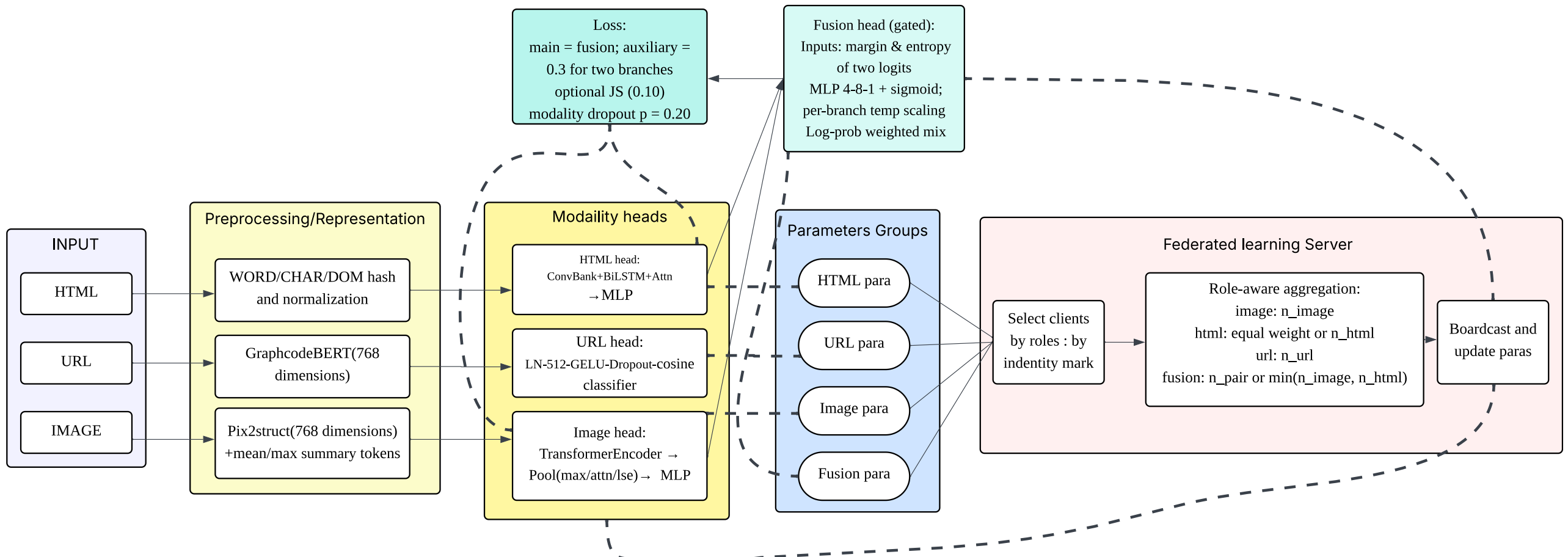

**Figure 1.** General workflow of Role-aware Federated-learning Phishing webpages detection

## 2.2 Environment and hyperparameters

Our experiments were conducted on the Artemis HPC system using the Slurm scheduler. We employed an Nvidia A-40 graphics card, 12 central processing units (CPUs), and 48 gigabytes (GB) of graphics processing unit (GPU) memory. The experimental environment was established based on Flower, as previously described, with 256 GB of memory (RAM). CUDA 11.3.1 was utilised to expedite the embedding and training processes. The training duration varied depending on different data partitioning strategies and experimental setups. It was also affected by the server throughput. Given that Artemis-HPC is a shared computing cluster, accurately controlling the training time was challenging. The number of federated aggregation rounds was set to 100. The number of local epochs was set to 5, the learning rate to 0.001, and the batch size to 64.

The maximum training time was approximately 16 hours, covering all data and all heads except the Fusion head, with 6 clients involved. The minimum training time was around 1 hour, using only URL data and with 2 clients. Moreover, due to substantial differences in loading speed across embedding methods, we also recorded the time intervals. The pix2struct embedding process took approximately 40 minutes. The static three-way split embedding for data from two HTML datasets took about 20 minutes, while the single-URL embedding using graphcodebert took about 5 minutes. The maximum RAM utilisation was approximately 100 GB, and the majority of computational resources and time were consumed by the IMAGE head and its associated components, accounting for roughly half of the total.

The four-head architecture and key hyperparameters are presented as follows:

1. **Image Head**: The image head utilised a TransformerEncoder. The input dimension was set to 768, the number of layers was 2, the number of attention heads was 8, the feed-forward dimension was 1024, and the dropout rate was 0.2. The pooling operation was governed by the global parameter POOL, configured as max pooling. The classification layer consisted of a LayerNorm layer, a full-connection (FC layer with 512 neurons, a Gaussian Error Linear Unit (GELU) activation function, a Dropout layer with a rate of 0.2, and an output layer.

2. **HTML Head**: The HTML head adopted a three-way lightweight encoding approach. The embedding dimensions for the character, word, and Document Object Model (DOM) branches were 64, 128, and 64, respectively. For the character branch, a multi-scale convolutional group with 16 kernels (ranging in size from 2 to 9) was employed, followed by a 128-dimensional FC layer. The word and DOM branches each incorporated a single-layer bidirectional Long Short-Term Memory (LSTM) network with a hidden dimension of 64, and attention pooling was applied. After concatenating the three branches, the same classification layer as the image head was used.
3. **URL Head**: URL Head. LayerNorm on 768-dim input; linear layer with 512 units (weight normalisation enabled), GELU, Dropout 0.2; then a cosine classifier to two classes. The classifier normalises features and class weights and uses a learnable scale initialised to 10. Key hyperparameters: hidden size 512, dropout 0.2, weight normalisation on or off, initial scale 10, number of classes 2.
4. **Fusion Head**: The fusion head was designed as a gated Multi-Layer Perceptron (MLP). Its input consisted of four statistical measures from two branches, namely the difference between the top 2 logits (margin) and the entropy of each branch. The architecture of the MLP was 4-8-1, with a Rectified Linear Unit (ReLU) activation function and a sigmoid output to generate the weight α. In the log-probability space, the two branches were weighted and combined according to α. Additionally, a learnable temperature parameter was used to calibrate each branch, with an initial value of 1.0.

For the outputs of the image head and HTML head, the same auxiliary loss was computed, with a coefficient of 0.30. This was used to stabilise the model and preserve the single-path capabilities, and could be toggled on or off. An optional Jensen-Shannon divergence term for consistency regularisation was included, with a coefficient of 0.10. Setting this coefficient to 0 effectively disabled this regularisation term. To enhance the model's robustness to missing modalities, during the pairwise training phase, one branch was randomly masked at the batch level with a probability of 0.20.

### 2.3 Aggregation algorithms design

In situations where there are fundamental differences in data modalities and types, directly applying federated averaging can give rise to parameter conflicts. These conflicts may prevent the model from aggregating properly or converging. Indeed, numerous studies have been conducted to address this issue [25][26][27]. Among them, the design of FedMM [27] has provided significant inspiration.

FedMM is tailored to handle modality heterogeneity and missing modalities. Instead of averaging all parameters across the entire network, it adopts an aggregation approach based on "modality splitting". Specifically, it only performs federated aggregation of feature extractors among clients sharing the same modality. Meanwhile, the classifiers are localised to avoid cross-site aggregation. Additionally, FedMM incorporates a regularisation term for aligning global modality prototypes. By using weights scheduled according to the training rounds, it smoothly transitions between the supervised loss and representation alignment. Empirical evidence has demonstrated that FedMM outperforms zero - padded Multi – FedAvg [38] and pure local training.

Under the same "modality-heterogeneous/missing-modality" federated setting, our study shares the same objective as FedMM: to enhance downstream classification through cross-site collaboration without sharing the original data, and to ensure that clients with only partial modalities can also benefit.

More specifically, FedMM does not train a unified multimodal fusion network. Instead, it federates single-modal feature extractors across different sites based on modality type. To avoid conflicts arising from structural inconsistencies, the classifiers of each client are localised. On the server side, for each modality, "model aggregation + global class prototypes" are maintained in parallel. During local training, dynamic weights are used to smoothly switch between the supervised Binary Cross-Entropy (BCE) loss and the prototype alignment loss (L2 distance to the prototypes). FedMM has reported superior performance compared to zero-padded Multi-FedAvg and pure local training on datasets such as TCGA-NSCLC/RCC, and its performance approaches the upper bound of centralised fusion. In contrast, our approach freezes the upstream encoders (for images and HTML/URL). We use pre-computed or overwritten embeddings as inputs. At the client side, we only optimise lightweight branch heads and gated Probability of Embedding (PoE) fusion heads (including learnable temperature parameters). To enhance the model's robustness to missing modalities, we employ techniques such as modality dropout, auxiliary losses, and an optional Jensen-Shannon (JS) divergence term for consistency regularisation. On the server side, we aggregate each branch head and fusion module based on parameter prefixes.

In summary, both our approach and FedMM adopt a "modality-based disassembly + privacy-preserving" federated collaboration strategy. However, the key difference lies in that FedMM focuses on "federating feature extractors at the representation layer and aligning prototypes". At the same time, our method emphasises "federating branch heads and fusion modules at the prediction layer". This design choice aims to reduce communication and computational overheads and simplify the deployment process. (It should be noted that FedMM requires backpropagation through feature extractors, which is the most computationally intensive part. The optimisation objective in the FedMM paper directly acts on the L2 distance between the extractor output and the prototype, along with the BCE term, which means that gradients are updated for the feature extractors.) Inspired by FedMM and integrating the conditional routing concept of the Mixture of Experts (MoE), we have designed a simple role-aware aggregation mechanism based on FedProx [39]. We partition the parameter subgraphs by "roles" (e.g., URL, HTML, or image). During communication, only the parameters corresponding to the activated roles are aggregated, while the rest remain local. This approach effectively avoids cross-modal gradient interference and ineffective averaging. The pseudo-code for this process is presented in Figures 2 and 3. (“shared” is a fallback mechanism when no acceptable weight is found, which will not affect the result.)

**Algorithm 1:** Role-aware FedProx (Server side)

**Input:** Rounds $R$; parameter name list
$\mathcal{N} = \text{keys}(\texttt{MultiModalNet.state_dict})$

**Data:** Global parameters $\theta$; **FedProx coefficient** $\mu$

1 **for** $r \leftarrow 1$ **to** $R$ **do**
`// line 2`
2 Broadcast $(\theta, \mu)$ to selected clients
3 Receive $\{(\theta_i, n_i, m_i)\}_{i \in \mathcal{I}_r}$ `//` $\theta_i$`: local params;` $n_i$`: samples;` $m_i$`: metrics (`$n$`.image/html/url, has.*)`
4 **foreach** $p \in \mathcal{N}$ **do**
5 $g \leftarrow \mathsf{GROUP}(p)$ `//` $g \in \{\text{image, html, url, fusion, shared}\}$
6 $S \leftarrow \mathsf{SELECT}(g, \{m_i\})$ `// only clients owning this role participate`
7 **if** $S = \varnothing$ **then**
8 $\theta^{\text{new}}[p] \leftarrow \theta[p]$ `// keep old if nobody can aggregate`
9 **else**
10 **if** $g =$ *image* **then**
11 $w_j \leftarrow m_j[n_{\text{image}}]$
12 **else if** $g =$ *html* **then**
13 $w_j \leftarrow 1$ `// equal weight; can switch to` $n_{\text{html}}$
14 **else if** $g =$ *url* **then**
15 $w_j \leftarrow m_j[n_{\text{url}}]$
16 **else if** $g =$ *fusion* **then**
17 $w_j \leftarrow m_j[n_{\text{pair}}]$ or $\min(m_j[n_{\text{image}}], m_j[n_{\text{html}}])$
18 **else**
19 $w_j \leftarrow n_j$ `// shared: sample-size weighting`
20 Normalize $w_j \leftarrow w_j / \sum_{k \in S} w_k$
21 $\theta^{\text{new}}[p] \leftarrow \sum_{j \in S} w_j \cdot \theta_j[p]$ `// role-wise weighted average`
22 $\theta \leftarrow \theta^{\text{new}}$; trigger ClientEvaluate and log per-client metrics (loss/accuracy/precision/recall/FPR)

**Figure 2.** Server-side algorithm flow for proposed Role-aware FedProx. (g refers to group; the client needs to provide an identity mark like “URL” before it participates)

**Algorithm 1:** Client-side Training with Paired Fusion (Dual-Head Updates)

**Input:** Local loaders $\mathcal{D}_{\mathrm{img}}, \mathcal{D}_{\mathrm{html}}, \mathcal{D}_{\mathrm{url}}$; paired loader $\mathcal{D}_{\mathrm{pair}}$; epochs $E$; lr $\eta$; focal loss $\mathcal{L}_{\mathrm{focal}}$; aux weight $\lambda_{\mathrm{aux}}$; JS weight $\lambda_{\mathrm{JS}}$; modal dropout $p$; clip $\tau$

**Data:** ImageHead, HtmlHead, FusionHead (with temperature scaling), UrlHead; snapshot $\theta^{(t)}$; FedProx $\mu$

1 **for** $e \leftarrow 1$ **to** $E$ **do**
2     **foreach** $(loader, mode) \in \{(\mathcal{D}_{\mathrm{img}}, \mathrm{image}), (\mathcal{D}_{\mathrm{html}}, \mathrm{html}), (\mathcal{D}_{\mathrm{url}}, \mathrm{url})\}$ **do**
3         **foreach** $(x, y) \in loader$ **do**
4             $o \leftarrow \mathrm{Model}(x, \mathrm{mode})$
5             $L \leftarrow \mathcal{L}_{\mathrm{focal}}(o, y) + \frac{\mu}{2} \sum_{p \in \mathcal{I}_{\mathrm{mode}}} \left\| \theta[p] - \theta^{(t)}[p] \right\|_2^2$    `// FedProx on this head`
6             backward $L$; clip to $\tau$; step
7     **if** $\mathcal{D}_{\mathrm{pair}}$ *exists* **then**
8         **foreach** $((x^{\mathrm{img}}, x^{\mathrm{html}}), y) \in \mathcal{D}_{\mathrm{pair}}$ **do**
9             $l_i \leftarrow \mathrm{ImageHead}(x^{\mathrm{img}})$; $\quad l_h \leftarrow \mathrm{HtmlHead}(x^{\mathrm{html}})$
10             sample $r \sim \mathcal{U}(0, 1)$; $\quad (l_i^{\star}, l_h^{\star}) \leftarrow (l_i, l_h)$    `// Batch-level modal dropout (robustness to missing modalities)`
11             **if** $r < p/2$ **then** $l_i^{\star} \leftarrow \perp$
12             **else if** $r < p$ **then** $l_h^{\star} \leftarrow \perp$
13             $l_f \leftarrow \mathrm{FusionHead}(l_i^{\star}, l_h^{\star})$
14             $L_f \leftarrow \mathcal{L}_{\mathrm{focal}}(l_f, y)$; $\quad L_i \leftarrow \mathcal{L}_{\mathrm{focal}}(l_i, y)$; $\quad L_h \leftarrow \mathcal{L}_{\mathrm{focal}}(l_h, y)$
15             **if** $\lambda_{\mathrm{JS}} > 0$ **then**
16                 $p_i \leftarrow \mathrm{softmax}(l_i), \quad p_h \leftarrow \mathrm{softmax}(l_h), \quad m \leftarrow \frac{1}{2}(p_i + p_h)$
17                 $JS \leftarrow \frac{1}{2}[\mathrm{KL}(p_i \| m) + \mathrm{KL}(p_h \| m)]$    `// JS consistency (teacher = mean of branches)`
18             **else**
19                 $JS \leftarrow 0$
20             $L \leftarrow L_f + \lambda_{\mathrm{aux}}(L_i + L_h) + \lambda_{\mathrm{JS}} \cdot JS + \frac{\mu}{2} \sum_{p \in \mathcal{I}_{\mathrm{fusion}}} \left\| \theta[p] - \theta^{(t)}[p] \right\|_2^2$    `// FedProx on fusion head`
21             backward $L$; clip to $\tau$; step
22     **if** $\mathcal{D}_{\mathrm{pair}}^{\mathrm{val}}$ *available* **then**
23         **foreach** $((x^{\mathrm{img}}, x^{\mathrm{html}}), y) \in \mathcal{D}_{\mathrm{pair}}^{\mathrm{val}}$ **do**
24             $l_i \leftarrow \mathrm{ImageHead}(x^{\mathrm{img}})$; $l_h \leftarrow \mathrm{HtmlHead}(x^{\mathrm{html}})$; $o \leftarrow \mathrm{FusionHead}(l_i, l_h)$
25             update metrics (loss, acc, prec/recall/FPR) with $(o, y)$
26     **else**
27         **foreach** $(x, y) \in \mathcal{D}_{\mathrm{img}}^{\mathrm{val}}$ **do**
28             update metrics with ImageHead$(x), y$
29         **foreach** $(x, y) \in \mathcal{D}_{\mathrm{html}}^{\mathrm{val}}$ **do**
30             update metrics with HtmlHead$(x), y$
31         **foreach** $(x, y) \in \mathcal{D}_{\mathrm{url}}^{\mathrm{val}}$ **do**
32             update metrics with UrlHead$(x), y$

**Figure 3.** Client-side algorithm flow for proposed Role-aware FedProx

We have also provided a basic explanation of the fundamental formula of FedProx. FedProx Formula:

$$w_{t+1}^{k} = \arg\min_{w} \left( f_k(w) + \frac{\mu}{2} |w - w_t|^2 \right) \tag{1}$$

$\boldsymbol{w_{t+1}^{k}}$**:** The optimised weight of the local model on client k after round t+1.
$\boldsymbol{f_k(w)}$ **:** The loss function for client k.
**μ :** The regularisation parameter (proximal term).
$\boldsymbol{w_t}$**:** The weight of the global model after round t.
$\boldsymbol{\arg\min_w}$: The argument of the minimum indicates that $w_{t+1}^{k}$ minimises the expression within the parentheses.

## 3 Dataset and preprocessing

### 3.1 Dataset selection

We selected two datasets that have been validated in previous studies. One is sourced from WebPhish, and the other is TR-OP, which was originally proposed by KnowPhish [7]. WebPhish consists of 45,373 balanced samples of HTML and URLs, and TR-OP contains 10,000 balanced samples.

The WebPhish dataset does not include images, whereas TR-OP includes images, original HTML, and short URLs corresponding to domain names, but without additional features. The HTML in WebPhish has undergone processing. It should be noted that the data complexity of the two datasets is entirely different, especially in terms of HTML data. Through our manual verification, the processed HTML in WebPhish is cleaner and shorter (with some truncation). In contrast, the HTML in TR-OP is more similar to the original HTML pages, containing substantial noise and irrelevant data, which makes it challenging to use directly for model training. We computed the maximum character length of the original text data in both datasets. To avoid biases stemming from word segmentation or word-list priors, we used pure character counting (len) for length statistics across all samples (note that this is for a rough statistical analysis only).

- For TR-OP's HTML, the global maximum character count is 79,647,959.
- For TR-OP's URL, the global maximum character count is 3,669.
- For WebPhish's HTML, the global maximum character count is 32,783.
- For WebPhish's URL, the global maximum character count is 2,624.

Regarding the reason why we did not utilise TR-OP's URLs, first, we considered the relatively small data volume. Additionally, URLs have fewer features and are shorter compared to HTML. Compared to the 45,373 URLs in WebPhish, TR-OP only has 10,000 URLs. Moreover, most of the URLs in TR-OP are closer to the original page URLs, without any suffixes, and carry minimal information, which may easily lead to an overestimation of performance.

### 3.2 Dataset samples and sources

The original data sources for the two datasets are OpenPhish [29] and PhishTank [30], and the harmless negative examples are from Tranco [31] and Alexa [32]. TR-OP corresponds to OpenPhish and Tranco. OpenPhish and PhishTank are well-known sources that provide automated detection and community-verified intelligence on phishing URLs. At the same time, Tranco offers a research-oriented, manipulation-resistant ranking of popular sites for harmless comparison. Alexa Top Sites provides a general ranking of popular websites to supplement the benign samples.

The original links of the two datasets are provided:

https://www.kaggle.com/datasets/guchiopara/look-before-you-leap (WebPhish)
https://github.com/imethanlee/KnowPhish?tab=readme-ov-file (TR-OP)

Currently, the data sources for the vast majority of phishing research exhibit similarities. Through our investigation of several pieces of literature and reviews [18][33][34], these open-source platforms have been verified by multiple parties and are generally considered credible.

However, a potential issue exists. As most studies have not accounted for data duplication and have all performed some degree of manual screening and pre-processing, it is extremely challenging to ensure there is absolutely no duplication. Meanwhile, due to processing and manual operations, there are also differences in data formats, as exemplified by Webphish and TR-OP.

To rule out overestimation of performance due to this factor, we included ablation experiments using a single dataset in the experimental section. This was done to verify the potential impact of data interconnection between the test and training sets.

### 3.3 Data preprocessing

Analogous to WebPhish and Web2vec, we formulate a lightweight, reproducible three-way HTML representation for each page: the character stream, the visible word stream, and the DOM tag stream. Initially, we conduct minimal normalisation solely on HTML.txt. This involves collapsing line breaks and tab characters into single spaces, consolidating multiple consecutive spaces, removing zero-width characters, and eliminating control characters that are disabled in Excel. Throughout this process, we refrain from altering any tags or attributes.) For the character stream, the complete HTML is encoded as UTF-8 bytes. It is then truncated or padded to a fixed length of CHAR_LEN = 4096. The values range from 0 to 255, with a padding value of PAD = 256. Regarding the visible word stream, we utilise BeautifulSoup to extract the text. Subsequently, Unicode-friendly tokenisation is carried out. This approach preserves letters, digits, combining marks, underscores, ZWJ (Zero-Width Joiner), and ZWNJ (Zero-Width Non-Joiner). Each token is mapped to one of the WORD_HASH = 131,071 buckets using FNV-1a 64-bit hashing. The sequence is then truncated and padded to a length of WORD_LEN = 1024, where the padding value is set to PAD = WORD_HASH. This strategy effectively circumvents issues related to large vocabularies and out-of-vocabulary (OOV) words.

For the DOM stream, we linearly traverse the tag names in the document tree. These tag names are hashed similarly to DOM_HASH = 8,190, and the resulting sequence is truncated and padded to a length of DOM_LEN = 1024, with PAD = DOM_HASH. This enables the capture of the page's structural patterns. All three outputs are represented as fixed-length LongTensor index sequences. Subsequently, they are embedded by independent lightweight encoders and integrated within the model. The entire preprocessing procedure exhibits linear-time complexity and requires only constant memory, making it inherently suitable for federated scenarios. Except for the optional sample shuffling, which uses a fixed random seed 42, the entire process is deterministic (specific parameters are detailed in Section 2.2). This characteristic not only facilitates peer review but also ensures the reproducibility of the experiments.

We selected a long sample from the processed TR-OP for demonstration, as shown in Table 1.

**Table 1 .** Processed HTML Sample example of TR-OP

| DOM_TOKEN | WORD_TOKEN | CHAR_TOKENS |
|---|---|---|
| html head meta script button div a img span iframe …… | login user name password cr360 cookies check want confirms my choices …… | </script><meta name="viewport" content="width=device-width, initial-scale=1"><style id="onetrust-style">…… |

We did not use GraphcodeBERT directly on HTML pages because BERT-based models have limited context length. HTML data, especially from TR-OP, varies greatly in length, often exceeding practical limits (see dataset description). Since BERT lacks a mechanism to select optimal tokens [35], we avoided using it for HTML embedding. Instead, we applied GraphcodeBERT to URLs, most of which are under the 512-token limit. For images, we preserved the original data without modification to avoid introducing bias and ensure fair comparison. The TR-OP dataset has already been manually curated in prior studies [11][12]. We used Pix2Struct [13], a pre-trained model designed for HTML and UI structures, to process image inputs. Specifically, we used the lightweight Pix2Struct-base. We added two global summary tokens at the start of the encoded sequence, generated via max pooling to capture overall image semantics and saliency (non-learnable). The image classifier is a basic Transformer Encoder followed by an MLP (parameters specified earlier).

# 4 Experimental results and analysis

## 4.1 Experimental steps and designs

Given the wide variety of data types and the simultaneous use of two datasets, to assess deviations in overall training performance across different client configurations, we have devised the following experimental protocols and client combinations. (Each set of experiments was carried out independently, and there were certain errors in the aggregated values, but the convergence should follow a similar trend):

1. **Six clients,** encompassing IMAGE, two distinct types of HTML, and URL concurrently.

2. **Four clients with diverse dual-data combinations,** specifically: the combination of IMAGE and HTML, WebPhish's HTML paired with URL, the pairing of Fusion and HTML, and HTML spanning across the two datasets.
3. **Two clients are designated for ablation experiments.** The aim is to validate the independent performance baselines of each head. These include clients solely containing HTML from one dataset, those with only IMAGE, those with just URL, and those with only Fusion

Through the above experimental design, we validated the framework's effectiveness under diverse data distributions. Each client was tested on unseen data; for example, if IMAGE was used for training, HTML served as the test set, and vice versa. When faced with significantly different test data, the model used heads trained on other clients for detection. The Fusion head was excluded from the six-client group because it artificially boosts the performance of the HTML and IMAGE heads (see Algorithm 2), making it impossible to evaluate their lower-bound performance in mixed configurations. Therefore, we included the Fusion head in the four-client group to demonstrate its enhancement effect.

We also considered non-IID scenarios. We conducted additional experiments by adjusting the training data distribution (e.g., reducing one client's HTML data from 1750 to 1000 and reallocating the rest) to assess the impact. Due to inherent structural differences between phishing web pages and images, non-IID conditions naturally arise, even under uniform distributions. For TR-OP, we selected data from indices 3500 to 5000 as the test set and shuffled the original order to avoid bias caused by time-based data collection. WebPhish data had already been shuffled in prior studies, so no further processing was applied. Hyperparameter settings are detailed in Section 2.2.

### 4.2 Results and analysis

Firstly, regarding the performance of all six clients, we separately documented the convergence curves and the final performance metrics (selecting one instance from five experiments. The performance metrics were uniformly set as Accuracy, False Positive Rate (FPR), Recall, and Precision). As depicted in Figure 4, in the case of no prox regularisation (with a coefficient of 0), we placed particular emphasis on documenting the curves of FPR and Accuracy (ACC).

The data was partitioned such that 3500 IMAGE samples from TR - OP and the HTML.txt file were designated as the training set. These were distributed among two clients, with each client receiving 1750 samples. The remaining samples were used as the test set and distributed to other clients who had not previously been exposed to HTML or IMAGE data. For WebPhish, 35000 samples of URLs and HTML samples were allocated to a single client as the training set, and the remaining samples were assigned to a client that had not encountered the corresponding samples as the test set (for example, a client that had been allocated the TR-OP image data).

This experimental design serves to validate the effect we initially proposed, namely, decoupling the training and test sets of clients. This allows any client to leverage the heads trained by other clients for detection purposes.

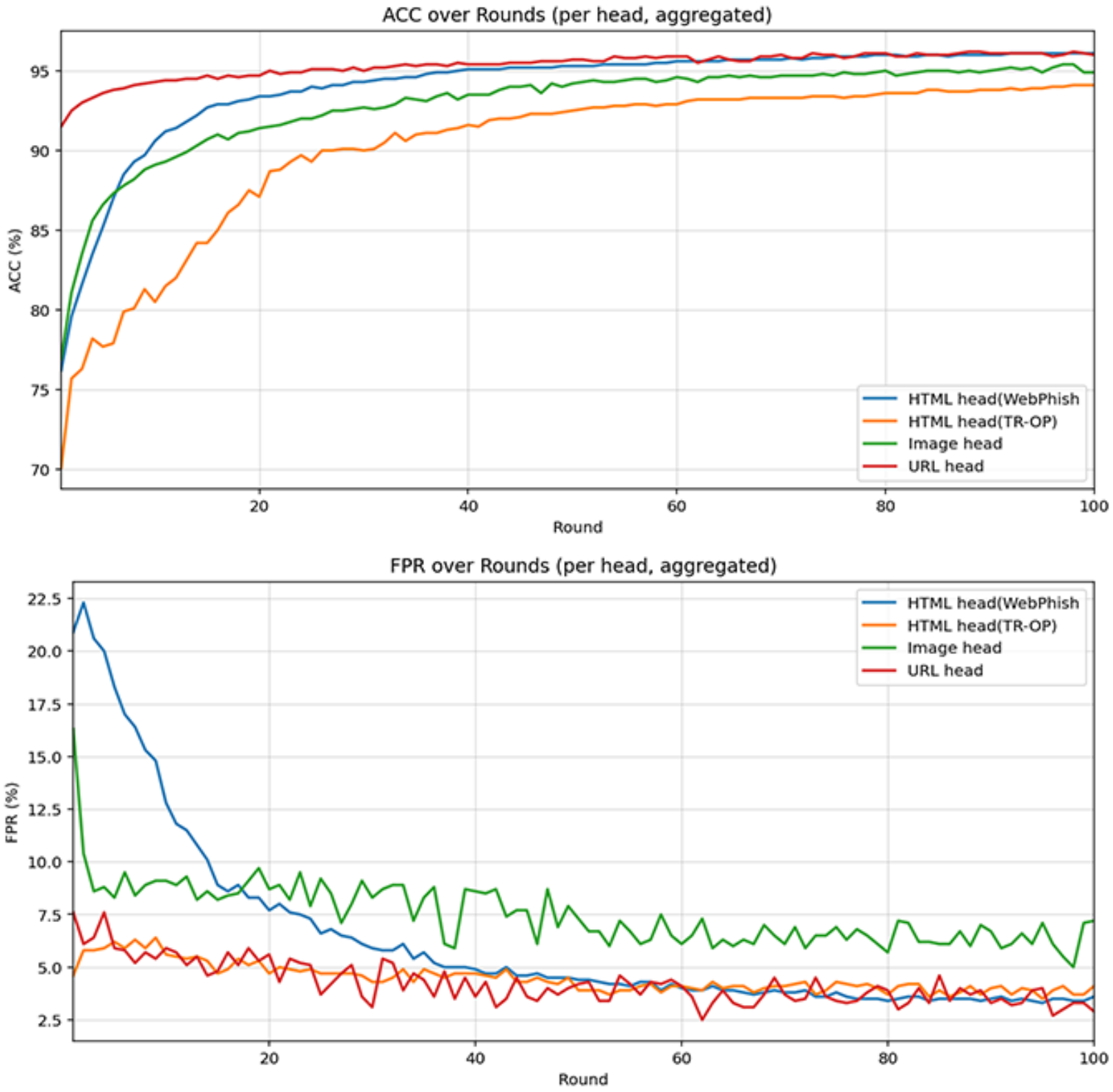

**Figure 4.** 6 clients' accuracy and FPR convergence curve over 100 rounds, includes HTML head, IMAEG head and URL head. Under fedavg.

Next, we analyse the four experimental groups. The data allocation follows the same method as in the six-client setup. After dividing the data into equal parts of 1750 samples per client, it is assigned to two clients. We selected the HTML-IMAGE and HTML-FUSION experiments as representative cases, shown in Figures 5 and 6. In cross-HTML testing, as mentioned earlier, we modified the partitioning of the train data for TR-OP HTML from 1750-1750 to 1000-2500 for 2 clients, and 35000 HTML of WebPhish is divided into 15000-20000 for the remaining 2 clients to simulate non-IID. We applied FedProx with a regularisation coefficient of 0.02, as illustrated in Figure 7. Since the HTML head weights are set equally, the large volume of WebPhish data slightly slows the convergence of the TR-OP client with limited data. However, this setup improves both the initial convergence speed and the False Positive Rate (FPR) performance on the small TR-OP test set. Moreover, when both HTML and Fusion clients coexist, the Fusion component enhances the performance of the HTML client, leading to gradual performance convergence.

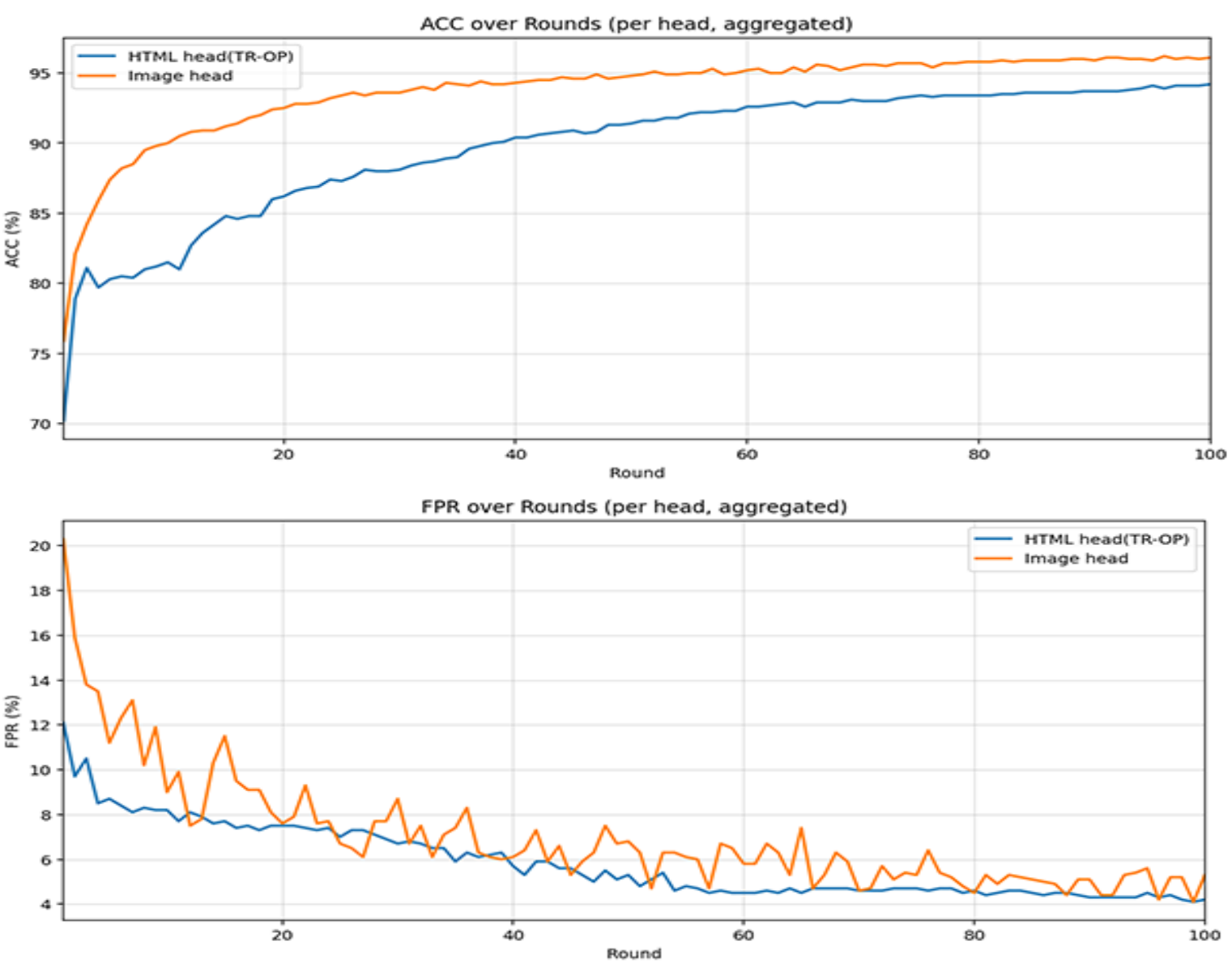

**Figure 5.** HTML-IMAGE heads convergence curve with 4clients and cross-dataset setting

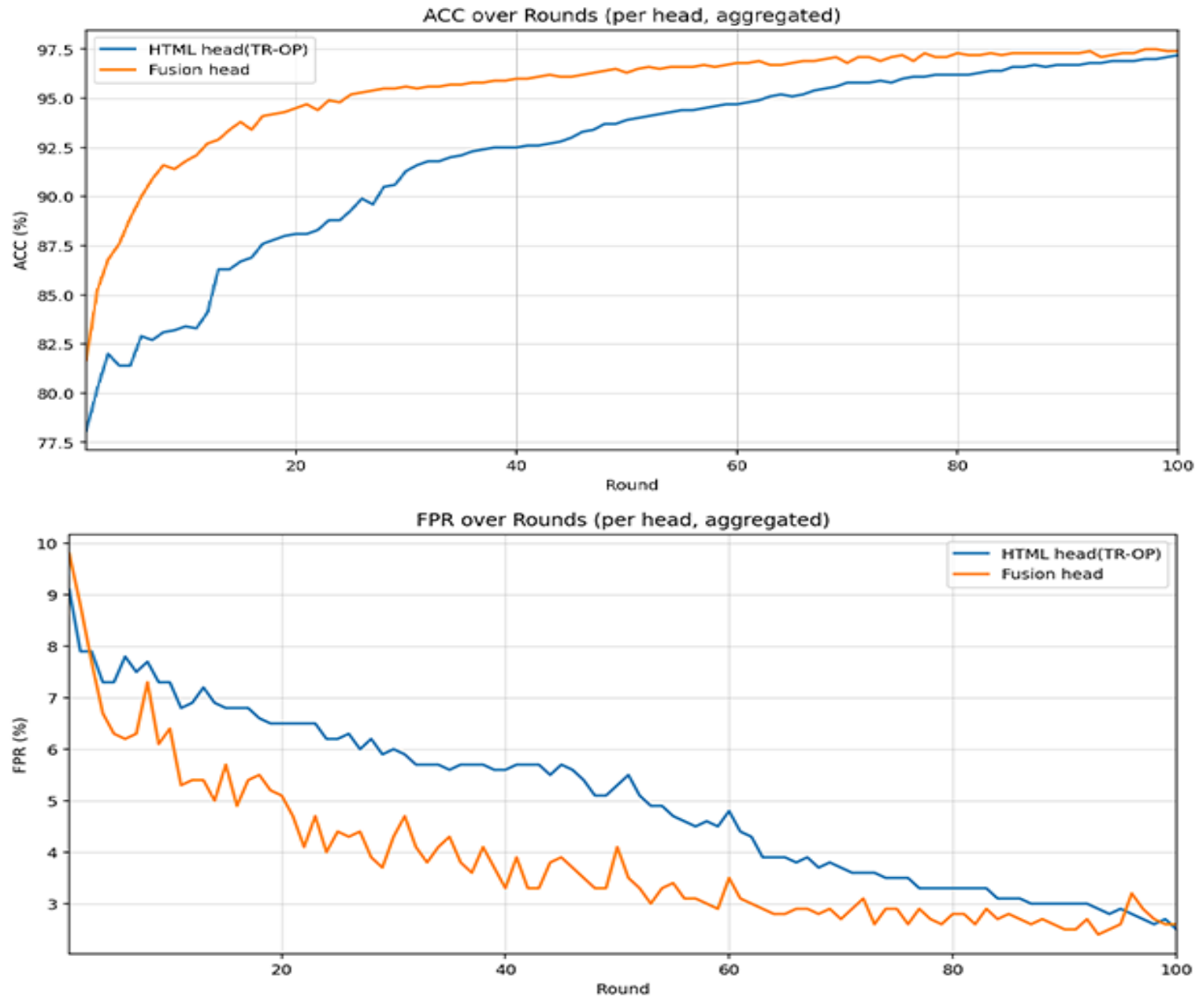

**Figure 6.** HTML-Fusion heads convergence curve with 4clients and cross-dataset setting

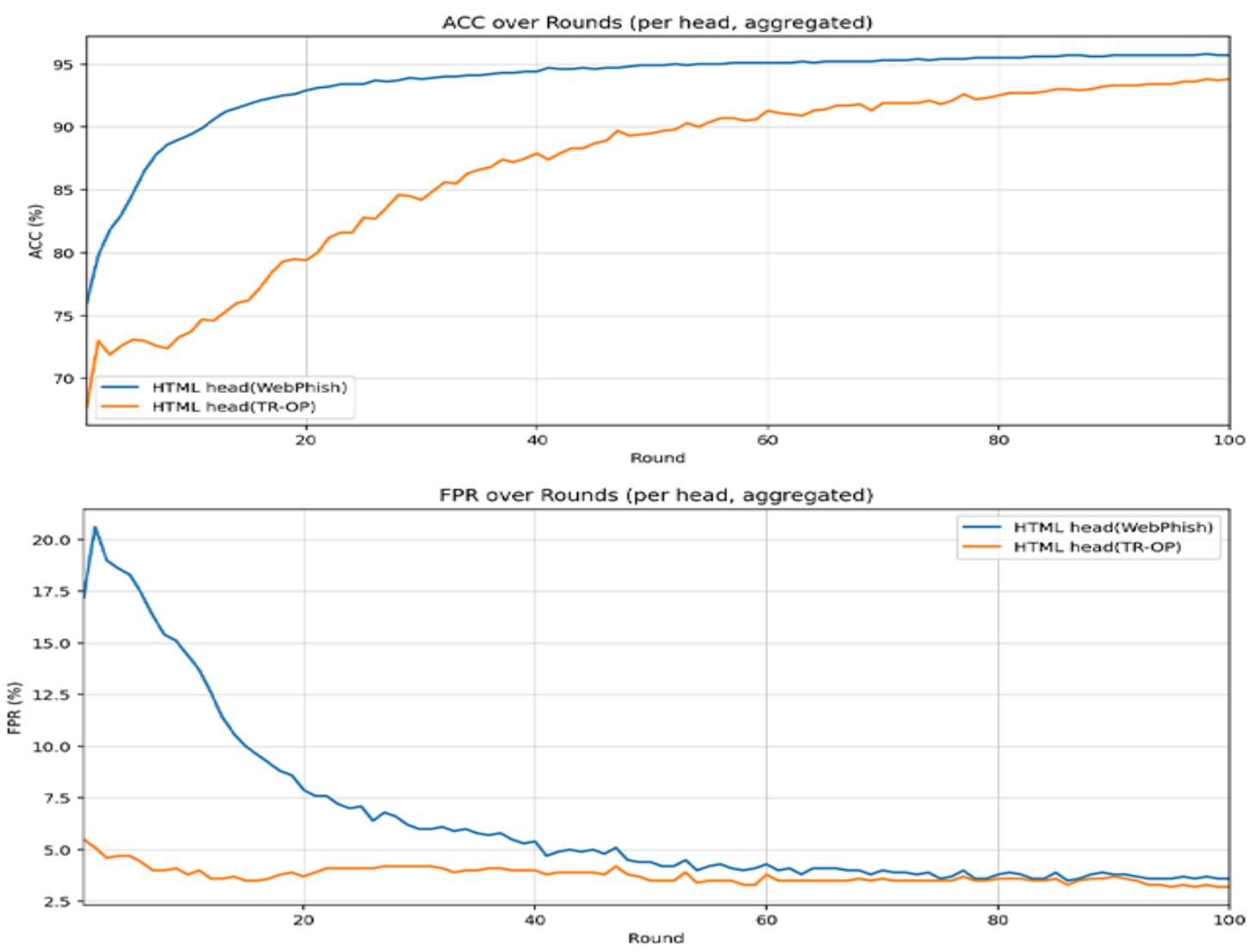

**Figure 7.** Convergence curves for HTML heads trained on 2 types of HTML datasets across 4 clients. (non-IID) , under fedprox = 0.02.

Finally, there are two client experimental groups for ablation tests. To verify the performance metrics that a single head can achieve in the absence of interference, in this experimental group, the test and training sets for the clients are of the same type, and the data is evenly distributed. See Figures 8 and 9.

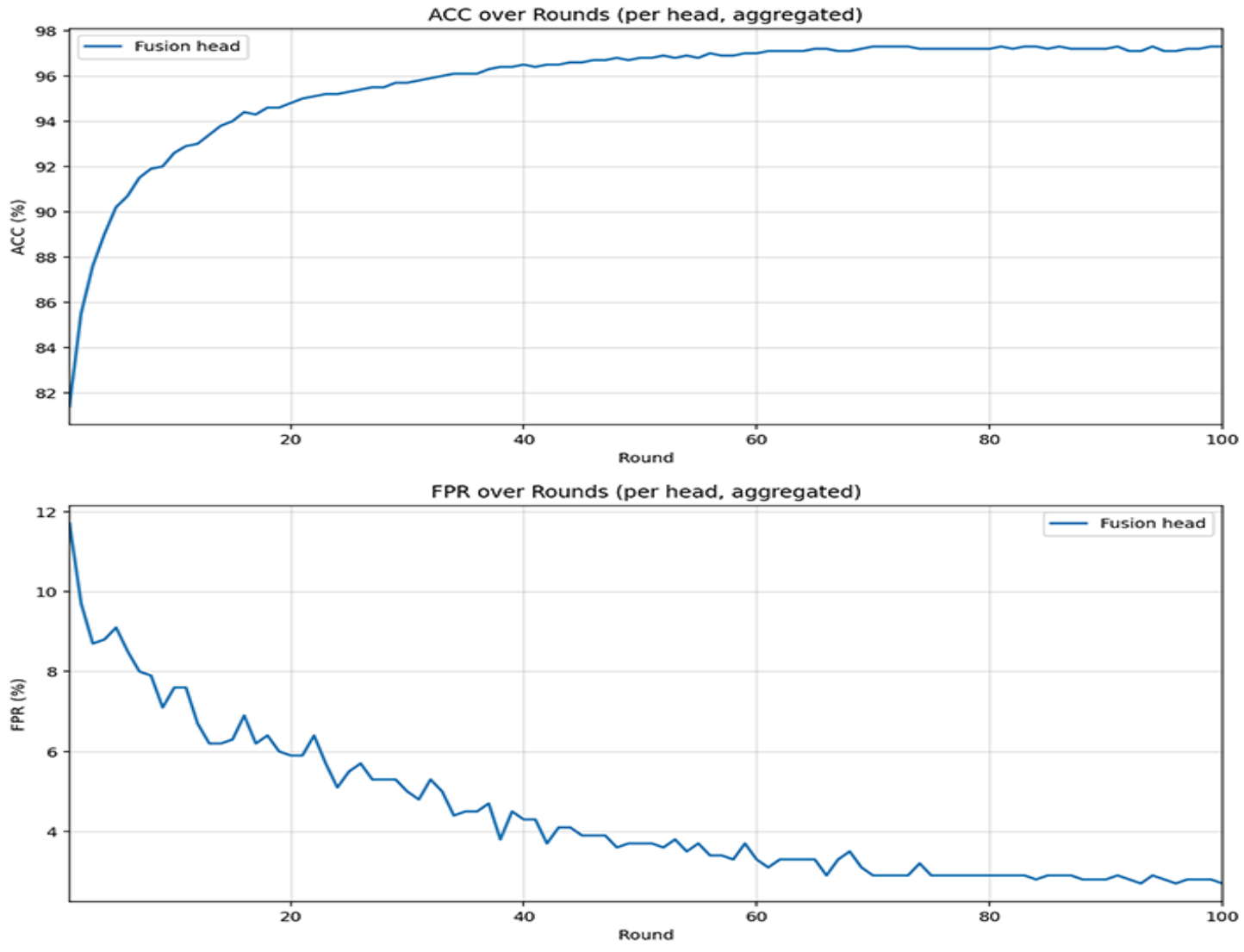

**Figure 8.** Fusion head convergence curve over 100 rounds with 2 clients trained on TR-OP.

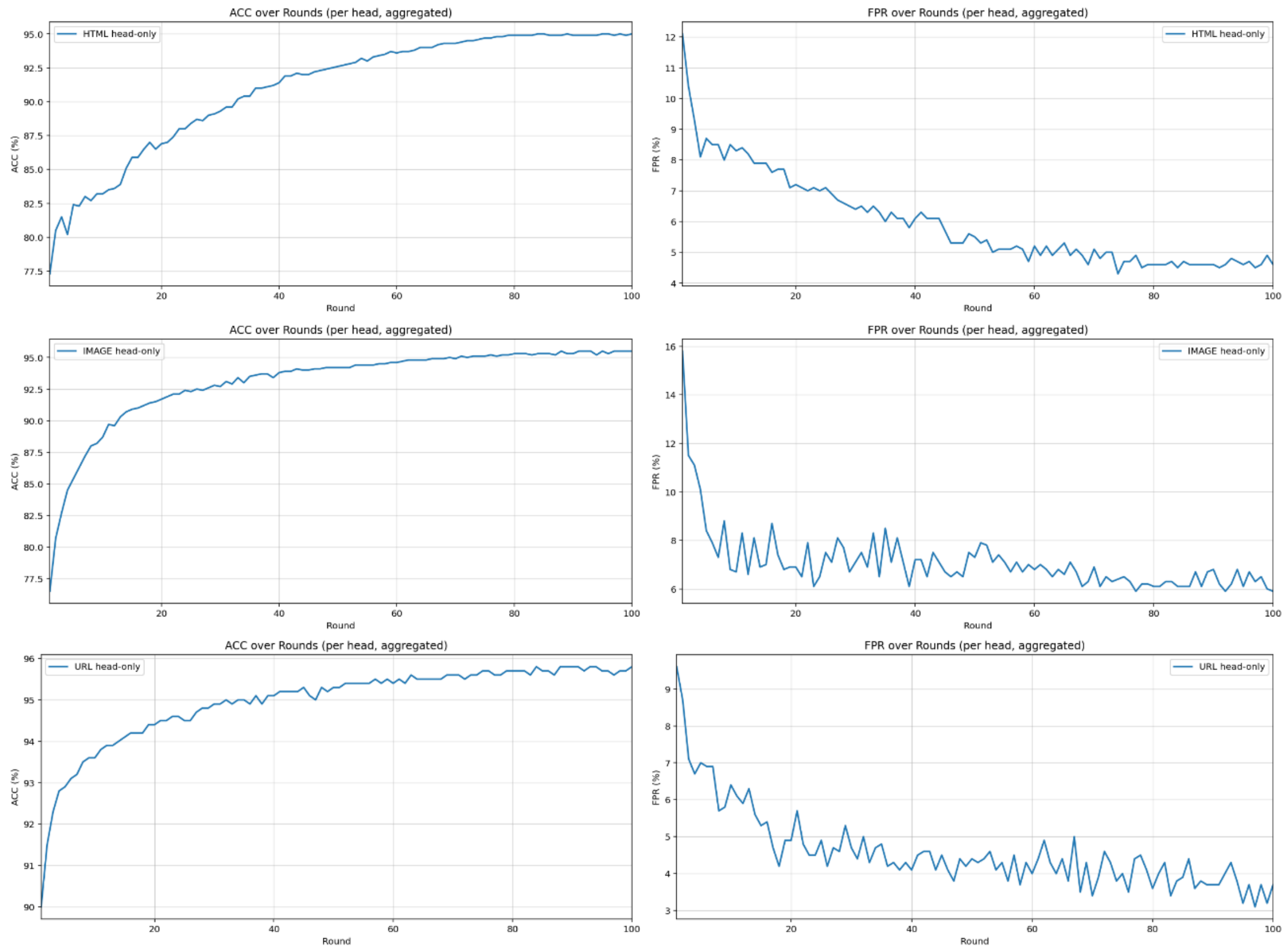

**Figure 9.** IMAGE, HTML, URL heads Accuracy and FPR convergence curve over 100 rounds when trained on 2 clients. (HTML heads trained and tested on TR-OP without WebPhish's influences, URL heads trained and tested on WebPhish)

We compared our method with two different schemes [8][11] that adopted the same dataset. Due to the averaging effect in federated learning, there was about a 1.5% difference in final performance across experiments. We selected the better performance among multiple experiments, as shown in Table 2. "Corresponding Test dataset" means whether the test set under the same dataset is needed. "Only" indicates the performance only under this type of data. "Final performance" refers to the comprehensive performance, which in our experiment is the FUSION head's performance. "NA" means "Not Applicable".

**Table 2 .** Baseline and comparison

| **Baseline** | **Methods** | **Corresponding Test dataset** | **Final Performance** | **Only-IMAGE** | **Only-HTML** | **Only-URL** |
|---|---|---|---|---|---|---|
| WebPhish [8] | Centralized End-to-End | True | Acc: 98.1%<br>Precision: 98.2%<br>Recall: 98.1% | NA | Acc: 96.5%<br>Precision: 97 %<br>Recall: 96.1% | Acc:95.9%<br>Precision: 95.6%<br>Recall:96.4% |
| PhishAgent [11] | Modularized Hybrid | Flexible | Acc: 96.1%<br>Precision:95.24%<br>Recall: 97.05% | Acc: 95.5%<br>Precision:95.27%<br>Recall: 95.75% | Acc: 85.88%<br>Precision: 84.18%<br>Recall: 95.75% | NA |
| **Our Method** | Privacy-preserving Federated learning | Flexible | Acc:97.5%<br>Precision:98.1%<br>Recall: 97.3%<br>FPR:2.4%<br>(On TR-OP) | Acc: 95.5%<br>Precision: 94.3%<br>Recall: **96.9%**<br>FPR:5.9%<br>(On TR-OP) | Acc: 96.5%<br>Precision: **98.1%**<br>Recall: 95.2%<br>FPR:2.6%<br>(on WebPhish) | Acc: 95.8%<br>Precision: **96.3%**<br>Recall: 95.3%<br>FPR:3.7%<br>(on WebPhish) |

## 5 Conclusion and future work.

This paper proposes a Role-Aware federated multimodal phishing detection framework based on FedProx. Through various combination experiments, we have verified the effectiveness of the entire framework under different data combinations. Meanwhile, we have designed a simple Role-Aware FedProx algorithm that utilises pre-labelling of input data to help the model distinguish different parameters and their corresponding aggregation paths, preventing parameter conflicts in federated learning. Our solution provides a new path and insights for jointly training a more flexible detection model while protecting privacy. However, the current design also has some limitations: 1. Under the current algorithm design, if both HTML and FUSION exist in the client, the HTML head or image head will be updated twice during backwards propagation. Although this speeds up the convergence of a single head, it also leads to double training and increased computational cost. We are considering how to optimise this process to reduce computational cost further while maintaining convergence speed. 2. The convergence effect of the image head on FPR is not stable. We believe this is related to the significant differences in the screenshots of phishing pages. We plan to use YOLO or other models to perform another round of cutting before inputting the data into the embedding model in the future, to collect key information points, such as input boxes and logos. However, this requires high precision and complete pre-training to ensure a clean cut, and it will also increase potential computational costs. We plan to further optimise the algorithm and model structure based on these issues in the future to reduce computational requirements further and improve overall performance.